\documentclass[10pt,twocolumn,letterpaper]{article}

\usepackage{cvpr}
\usepackage{times}
\usepackage{epsfig}
\usepackage{graphicx}
\usepackage{amsmath}
\usepackage{amssymb}

\usepackage[pagebackref=true,breaklinks=true,letterpaper=true,colorlinks,bookmarks=false]{hyperref}

\cvprfinalcopy 

\def\cvprPaperID{1924} 

\ifcvprfinal\fi
\begin{document}

\title{Motion-Appearance Co-Memory Networks for Video Question Answering}

\author{Jiyang Gao$^{*}$ \quad Runzhou Ge$^{*}$  \quad Kan Chen \quad Ram Nevatia \\
University of Southern California \\
{\tt\small \{jiyangga, rge, kanchen, nevatia\}@usc.edu} 
}

\maketitle

\newcommand\blfootnote[1]{%
  \begingroup
  \renewcommand\thefootnote{}\footnote{#1}%
  \addtocounter{footnote}{-1}%
  \endgroup
}

\blfootnote{$*$ indicates equal contributions.}



\begin{abstract}
Video Question Answering (QA) is an important task in understanding video temporal structure. We observe that there are three unique attributes of video QA compared with image QA: (1) it deals with long sequences of images containing richer information not only in quantity but also in variety; (2) motion and appearance information are usually correlated with each other and able to provide useful attention cues to the other; (3) different questions require different number of frames to infer the answer. Based on these observations, we propose a motion-appearance co-memory network for video QA. 
Our networks are built on concepts from Dynamic Memory Network (DMN) and introduces new mechanisms for video QA. Specifically, there are three salient aspects: (1) a co-memory attention mechanism that utilizes cues from both motion and appearance to generate attention; (2) a temporal conv-deconv network to generate multi-level contextual facts; (3) a dynamic fact ensemble method to construct temporal representation dynamically for different questions. We evaluate our method on TGIF-QA dataset, and the results outperform state-of-the-art significantly on all four tasks of TGIF-QA.
\end{abstract}


\section{Introduction}
\label{sec:intro}

Understanding video temporal structure is an important topic in computer vision. To achieve this goal, various tasks have been proposed, such as temporal action localization \cite{Shou_2016_CVPR, Gao_2017_cbr}, action anticipation \cite{gao2017red} and video prediction \cite{villegas2016decomposing}. Besides these tasks, video Question Answering (QA) \cite{Jang_2017_CVPR, Tapaswi_2016_CVPR} is another challenging task, which not only requires the understanding of video temporal structure, but also joint reasoning of videos and texts. In this paper, we tackle the problem of video QA.

Image and text question answering have achieved much progress recently. The success comes in part from the application of attention mechanisms \cite{san_Yang_2016_CVPR, hie_coatten_lu2016hierarchical} and memory mechanisms \cite{kumar2016ask} in deep neural networks. Attention mechanisms tell the neural network ``where to look", while the memory mechanism refines answers in multiple reasoning cycles. Video QA is different from image QA \cite{Malinowski_2015_ICCV, hie_coatten_lu2016hierarchical} in two aspects:  (1) the questions are more about temporal reasoning of the videos, \eg motion transition and action counting, than spatial attributes, such as colors, spatial locations, which require effective temporal representation modeling; (2) the input source is a sequence of images, rather than a single image, which contains richer information not only in quantity but also in variety (appearance, motion, transition) to ``remember", and it makes the reasoning process more complicated.


Dynamic Memory Networks (DMN) \cite{kumar2016ask, xiong2016dynamic} were originally proposed for text and image question answering. It contained a memory module to encode the input sources multiple cycles and an attention mechanism allowing the reading process to focus on different contents in each cycle. Although DMN contains an input module and a memory module which are able to read and remember a long sequence information, which is applicable for videos, directly applying such a method to video QA task would not give satisfying results. Because it lacks motion analysis, especially joint analysis between motion and appearance in videos, and temporal modeling. To strengthen the memory mechanism, Na \etal \cite{Na_2017_ICCV} proposed a read-write memory network that jointly encode the movie appearance and caption content, however it lacks motion analysis and dynamic memory update. Xu \etal \cite{xu_2017_video} exploited the appearance and motion via gradually refined attention, where the motion and appearance features are fused together. 

We observe two unique attributes of answering questions in videos. The first is that the motion and appearance information are usually correlated with each other in the reasoning process. For example, in answering the question ``what does the woman do after look uncertain?" as shown in Figure \ref{fig:problem}, we need to first localize ``the woman look uncertain" action, which requires motion evidence for looking uncertain and appearance evidence for the woman; after that, we need to ignore the man's interval, and then focus on what the woman does (smile). Appearance and motion information are both involved in the reasoning process and provide attention cues to each other. The second attribute is that different types of questions may require representations from different amounts of frames, for example, ``what is the color of the bulldog?" needs only a single frame to produce the answer, while ``How many times does the cat lick" needs the understanding of the whole video.

Based on these observations, we propose a motion-appearance co-memory network for video QA. Our model is built on concepts of DMN/DMN+ \cite{kumar2016ask, xiong2016dynamic}, so we share the same terms with DMN \cite{kumar2016ask}, such as facts, memory and attention. Specifically, a video is converted to a sequence of motion and appearance features by the two-stream models \cite{xiong2016cuhk}. The motion and appearance features are then fed into a temporal convolutional and deconvolutional neural network to build multi-level \textit{contextual facts}, which have the same temporal resolution but represent different contextual information. These contextual facts are used as input \textit{facts} to the memory networks. The co-memory networks hold two separate memory states, one for motion and one for appearance. To jointly model and interact with the motion and appearance information, we design a co-memory attention mechanism that takes motion cues for appearance attention generation, and appearance cues for motion attention generation. Based on these attentions, we design \textit{dynamic fact ensemble} method to produce temporal facts dynamically at each cycle of fact encoding. We evaluate our model on TGIF-QA dataset \cite{Jang_2017_CVPR}, and outperform state-of-the-art performance significantly on all four tasks in TGIF-QA.

The novelty of our method is three-fold compared with DMN/DMN+ \cite{kumar2016ask, xiong2016dynamic}:

(1) We design a co-memory attention mechanism to jointly model motion and appearance information.

(2) We use temporal conv-deconv networks to build multi-level contextual facts for video QA.

(3) We introduce a method called dynamic fact ensemble to dynamically produce temporal facts in each cycle of fact encoding.

In the following, we first introduce related work, and then outline the DMN/DMN+ framework. In Section \ref{sec:method}, we present our motion-appearance co-memory network in detail, and in Section \ref{sec:eval}, we show the evaluation of our method on TGIF-QA.  

\section{Related Work}

\textbf{Image question answering.} 
Image question answering aims to measure the capability of reasoning about linguistic and image inputs jointly. Many methods have been proposed \cite{san_Yang_2016_CVPR, chen2015abc_cnn, fda_ilievski2016focused, hie_coatten_lu2016hierarchical, xu2016ask, fukui2016multimodal, Antol_2015_ICCV, Andreas_2016_CVPR, xiong2016dynamic, Malinowski_2015_ICCV, Kanchen_2018_CVPR, shih2016look, kim2016multimodal, Hu_2017_ICCV, ZhuChen_2017_ICCV, yu2017multi, Yuke_2017_ICCV, zhu2016visual7w}. 
Among all these models, attention mechanism \cite{san_Yang_2016_CVPR, chen2015abc_cnn, hie_coatten_lu2016hierarchical, zhu2016visual7w} provides guidance to deep models on ``where to look'' and  memory mechanism \cite{kumar2016ask, xiong2016dynamic} allows the model to have multiple reasoning iterations and refine the answer gradually. 
Question-guided attention mechanism \cite{chen2015abc_cnn} uses semantic representation of a question as query to search for the regions in an image that are related to the answer. Yang \etal~\cite{san_Yang_2016_CVPR} presented a Stacked Attention Network (SAN) that queries an image multiple times to infer the answer progressively. 
Lu \etal~\cite{hie_coatten_lu2016hierarchical} argued that modeling ``what words to listen to'' is equally important to model ``where to look'', and proposed a co-attention model that jointly reasons about image-guided and question-guided attention. Instead of directly inferring answers from the abstract visual features, Yu \etal~\cite{yu2017multi} developed a semantic attention mechanism to select high-level question-related concepts. Dynamic memory network (DMN), which was first introduced by Kumar \etal~\cite{kumar2016ask} to solve text based question answering, adopted episodic memories and attention mechanisms which allow multiple cycles of reasoning. Xiong \etal~\cite{xiong2016dynamic} improved the memory and input module of DMN so that it can be applied to image QA.

\textbf{Video question answering.} Video QA is a relatively new task compared with image QA.
Yu \etal~\cite{Yu_2017_CVPR} adopted a semantic attention mechanism, which combines the detected concepts in videos with text
encoding/decoding to generate answers.
Comparing with images, temporal domain is unique to videos. A temporal attention mechanism is leveraged to selectively attend to one or more periods of a video in \cite{Jang_2017_CVPR, Mun_2017_ICCV, xu_2017_video}.
Besides temporal attention mechanism, Jang \etal~\cite{Jang_2017_CVPR} and Xu \etal~\cite{xu_2017_video} also utilized motion information along with appearance information in videos. 
Recently Na \etal~\cite{Na_2017_ICCV} and Kim \etal~\cite{kim_2017_deepstory} both introduced the memory mechanism to their models for video QA. However, their models \cite{Na_2017_ICCV, kim_2017_deepstory} both lack motion analysis and dynamic memory update mechanism.

\textbf{Video temporal analysis.} To answer the video-based questions correctly, temporal analysis of videos is necessary.
Shou \etal~\cite{Shou_2016_CVPR} presented a multi-stage Segment-CNN model to generate action proposals and localize actions in videos.
Temporal Unit Regression Network (TURN) \cite{Gao_2017_ICCV_TURN} and Cascaded Boundary Regression (CBR) \cite{Gao_2017_cbr} exploit the temporal boundary regression mechanism for proposal generation and action detection. 
Recently Gao \etal~\cite{Gao_2017_ICCV_TALL} and Hendricks \etal~\cite{Hendricks_2017_ICCV} proposed to localize activities by language queries, their methods involve of joint modeling of the videos and language queries, which also related to video QA.

\section{General Dynamic Memory Networks}

\begin{figure}[]
  \centering
    \includegraphics[width=0.46\textwidth]{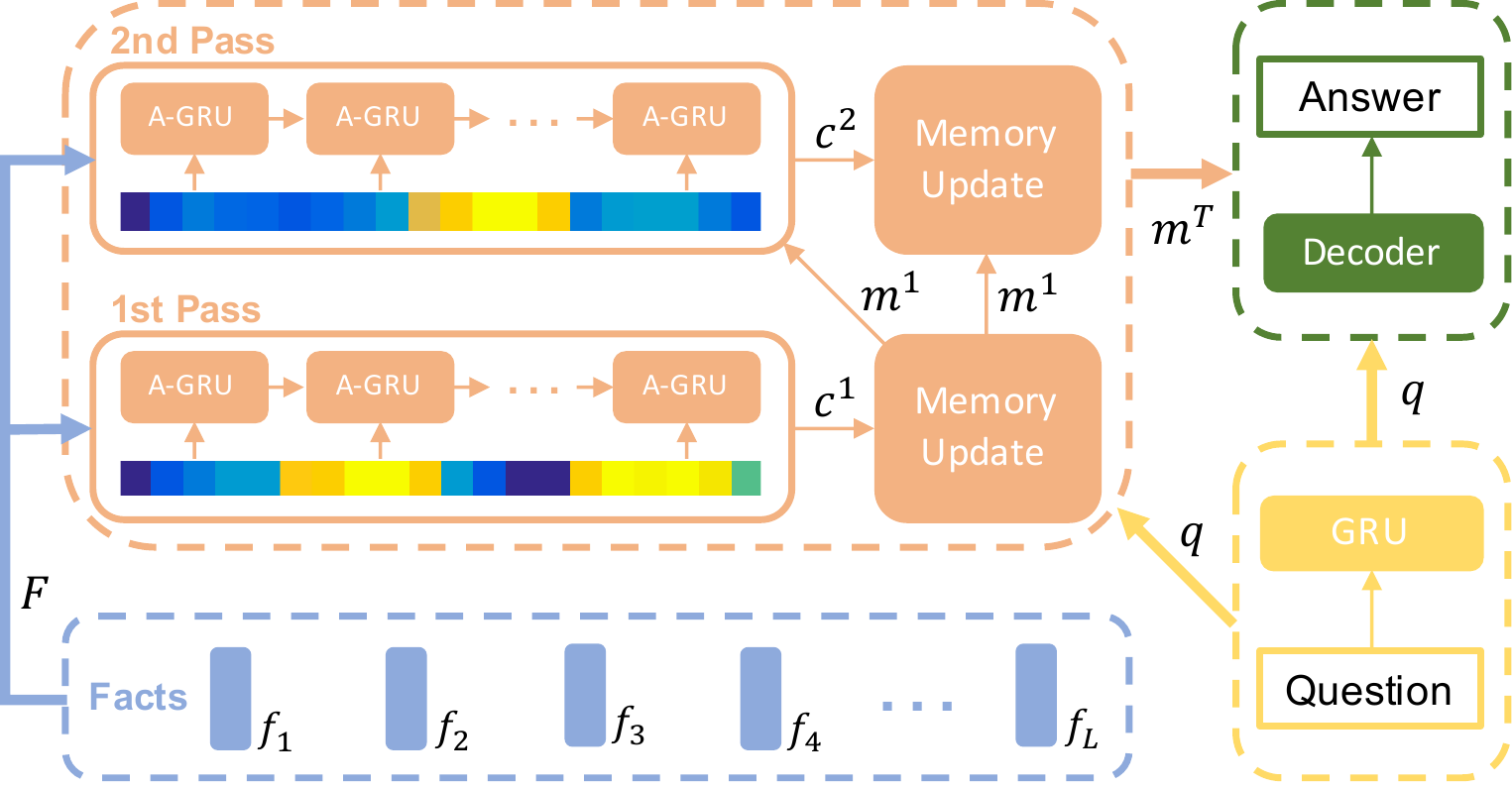}
    \caption{General Dynamic Memory Network (DMN) \cite{kumar2016ask} architecture. The memory update process for the $t$-th cycle is : (1) the facts $F$ are encoded by an attention-based GRU in episodic memory module, where the attention is generated by last memory $m^{t-1}$; (2) the final hidden state of the GRU is called contextual vector $c^t$, which is used to update the memory $m^t$ together with question embedding $q$. The question answer is generated from the final memory state $m^T$.}
      \label{fig:general_dmn}
\end{figure}

As our work is closely related to DMN \cite{kumar2016ask, xiong2016dynamic}, we begin with introducing the general framework of DMN. It contains four distinct modules: an input module, a question module, an episodic memory module and an answer module, as shown in Figure \ref{fig:general_dmn}. 

\textbf{Fact module.} The fact module converts the input data (\eg text, image, video) into a set of vectors called \emph{facts}, which is denoted as $F = [f_1, f_2,...,f_{L}]$, where $L$ is the total number of facts. For text-based QA, \cite{kumar2016ask} used a Gated Recurrent Unit (GRU) to encode all text information; for image-based QA, \cite{xiong2016dynamic} adopted a bi-directional GRU to encode the local region visual features to globally-aware facts.

\textbf{Question module.} The question module converts the question into an embedding $q$. Specifically, \cite{kumar2016ask, xiong2016dynamic}  used a GRU to encode the question sentence and use the final hidden state of the GRU as the question embedding.

\textbf{Episodic memory module.} Episodic memory is designed to retrieve the relevant information from the facts. To extract information related to the questions from the facts more effectively, especially when transitive reasoning is required in questions, the episodic memory module iterates over the input facts for multiple cycles, and updates the memory after each cycle. There are two important mechanisms in the episodic memory module: an attention mechanism and a memory update mechanism. 

Suppose that the updated memory after $t$-th cycle is $m^t$, the facts set $F=[f_1,f_2,...,f_{L}]$, the question embedding is $q$, then the attention gate $g_i^t$ is given by

\begin{equation}
    g_i^t=F_a(f_i,m_{t-1},q) 
\end{equation}    
where $F_a$ is an attention function which takes the fact vector $f_i$ at step $i$, memory $m^{t-1}$ at cycle $t-1$ and the question $q$ as inputs, and outputs a scalar value $g_i^t$, which represents the attention value for the fact $f_i$ in cycle $t$.

To effectively use the ordering and positional information in videos, an attention based GRU is designed. Instead of using the original update gate in the GRU, the attention gate $g_i^t$ is used, the update equation for the modified GRU is 
\begin{equation}
    h_i=g_i^t\circ\tilde{h_i}+(1-g_i^t)\circ h_{i-1}
\end{equation}
The final hidden state of the attention based GRU is used as the contextual feature $c^t$ for updating the episodic memory $m^t$. Together with the question embedding $q$ and the memory for cycle $t-1$, the $t$-th cycle memory is updated by 
\begin{equation}
    m^t=F_m(m^{t-1}, c^t, q)
\end{equation}
where $F_m$ is a memory update function. The final memory $m^T$ is passed to the answer module to generate the final answers, where $T$ is the number of memory update cycle.

\textbf{Answer module.} The answer module takes both $q$ and $m^T$ to generate the model’s predicted answer. Different answer decoders may be applied for different tasks, \eg a softmax output layer for single word answer.

\section{Motion-Appearance Co-Memory Networks}

\label{sec:method}
In this section, we present our motion-appearance co-memory networks, including multi-level contextual facts,  co-memory module and answer module. The question module remains the same as the one in traditional DMN.

\subsection{Multi-level Contextual Facts}
\label{sec:facts}
The videos are cut into small units \cite{Gao_2017_ICCV_TURN} (a sequence of frames). For each video unit, we use two-stream CNN models \cite{xiong2016cuhk} to extract unit-level motion and appearance features. More feature pre-processing details are given in Section \ref{sec:eval}. The sequence of unit-level appearance features and motion features is represented as $\{a_{i}\}$ and $\{b_{i}\}$ respectively. 

\begin{figure}[]
  \centering
    \includegraphics[width=0.40\textwidth]{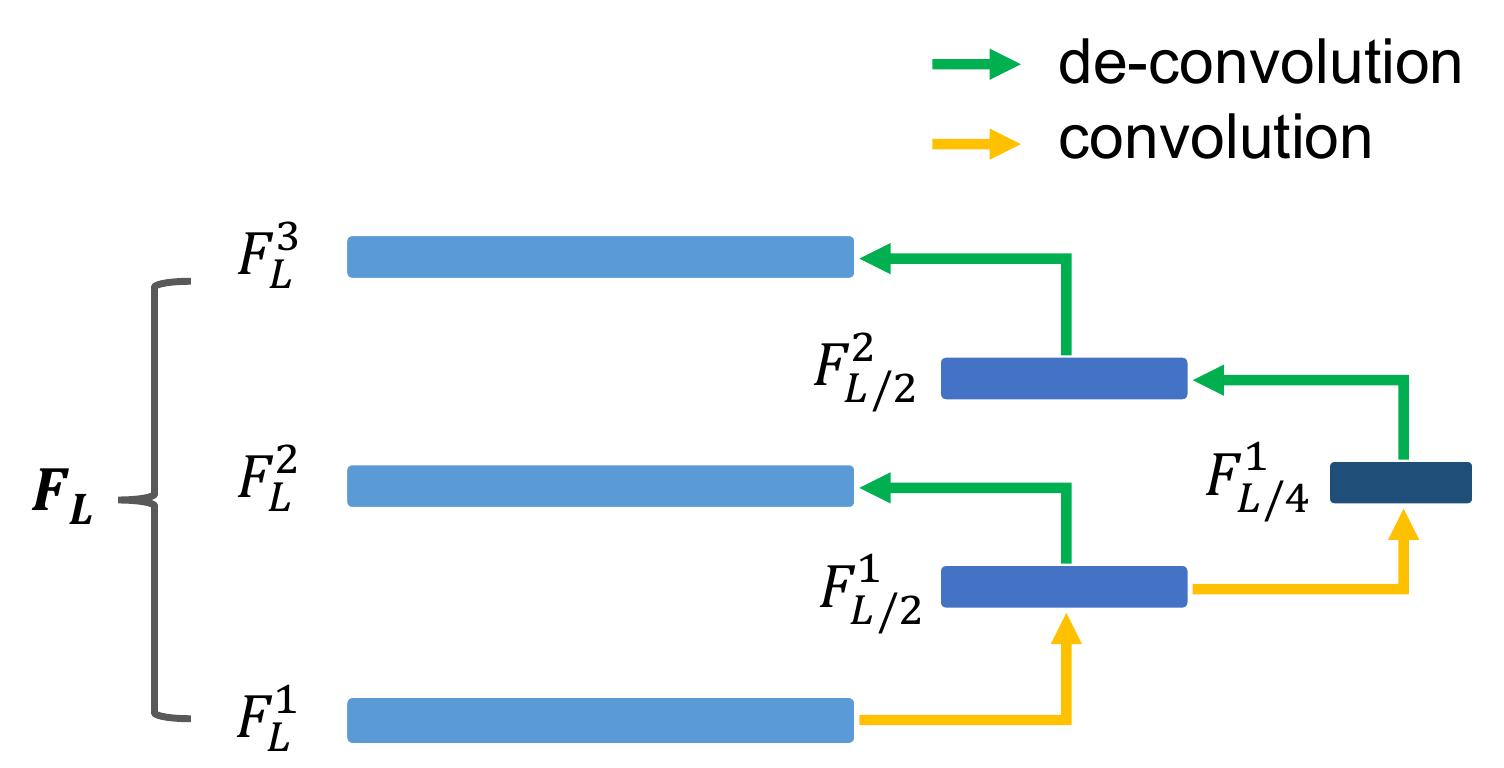}
    \caption{The input temporal representations are processed by temporal conv-deconv layers to build multi-layer contextual facts, which have the same temporal resolution but different contextual information.}
      \label{fig:facts}
\end{figure}

To build multiple levels of temporal representations where each level represent different contextual information, we use the temporal convolutional layers to model the temporal contextual information and de-convolutional layers to recover temporal resolution, as shown in Figure \ref{fig:facts}. Specifically, the lowest level feature sequence is built directly from the unit features, $A^1_{L}=\{a_{i}\}$, $B^1_{L}=\{b_{i}\}$. The convolutional layers compute a feature hierarchy consisting of temporal feature sequences at several scales with a scaling step of 2, $F^1_{L}$, $F^2_{L/2}$, $F^3_{L/4}$, ..., as shown in Figure \ref{fig:facts}. Note that $F$ could be $A$ (for appearance features) or $B$ (for motion features). The de-convolutional pathway hypothesizes higher resolution features $F^2_{L}$, $F^3_{L}$ by upsampling temporally coarser, but semantically stronger, feature sequences. Thus, $F^1_{L}$, $F^2_{L}$ and $F^3_{L}$ have the same resolution but different temporal contextual coverage. Note that we only show 3 levels in Figure \ref{fig:facts}, more levels could be modeled by adding more convolutional and de-convolutional layers. $\mathbf F_L=\{F^1_{L}, F^2_{L}, ... , F^N_{L}\}$ is termed as \emph{contextual facts}.

\subsection{Motion-appearance Co-Memory Module}
In this part, we introduce the co-memory attention mechanism and the dynamic fact ensemble method.

\textbf{Co-memory attention.} The questions in video QA usually involve both appearance and motion. 
Appearance usually provides useful cues for motion attention, \ie guides the focus on motion content, and vice versa. To allow interaction between appearance and motion,
we design a co-memory attention mechanism. Specifically, two separate memory modules are used to hold motion memory $m_b^t$ and appearance memory $m_a^t$, where $t$ is the number of cycle for memory update. As indicated before, when the networks read motion facts to update motion memory, appearance memory provides useful cues to generate attentions; motion memory is also helpful for updating appearance attention. Therefore, $m_b^{t-1}$ and $m_a^{t-1}$ are both used to generate attentions for motion and appearance fact encoding in the $t$-th cycle. As we build multiple levels of facts, we generate an attention score for each fact vector at each level. The motion attention gate for fact $b_j^i$  is $gb_{i,j}^t$ and the appearance attention for fact  $a_j^i$ is $ga_{i,j}^t$, where $t$ means the number of cycle, $i$ is the level of fact representation and $j$ is the step of the facts.
\begin{figure}[]
  \centering
    \includegraphics[width=0.48\textwidth]{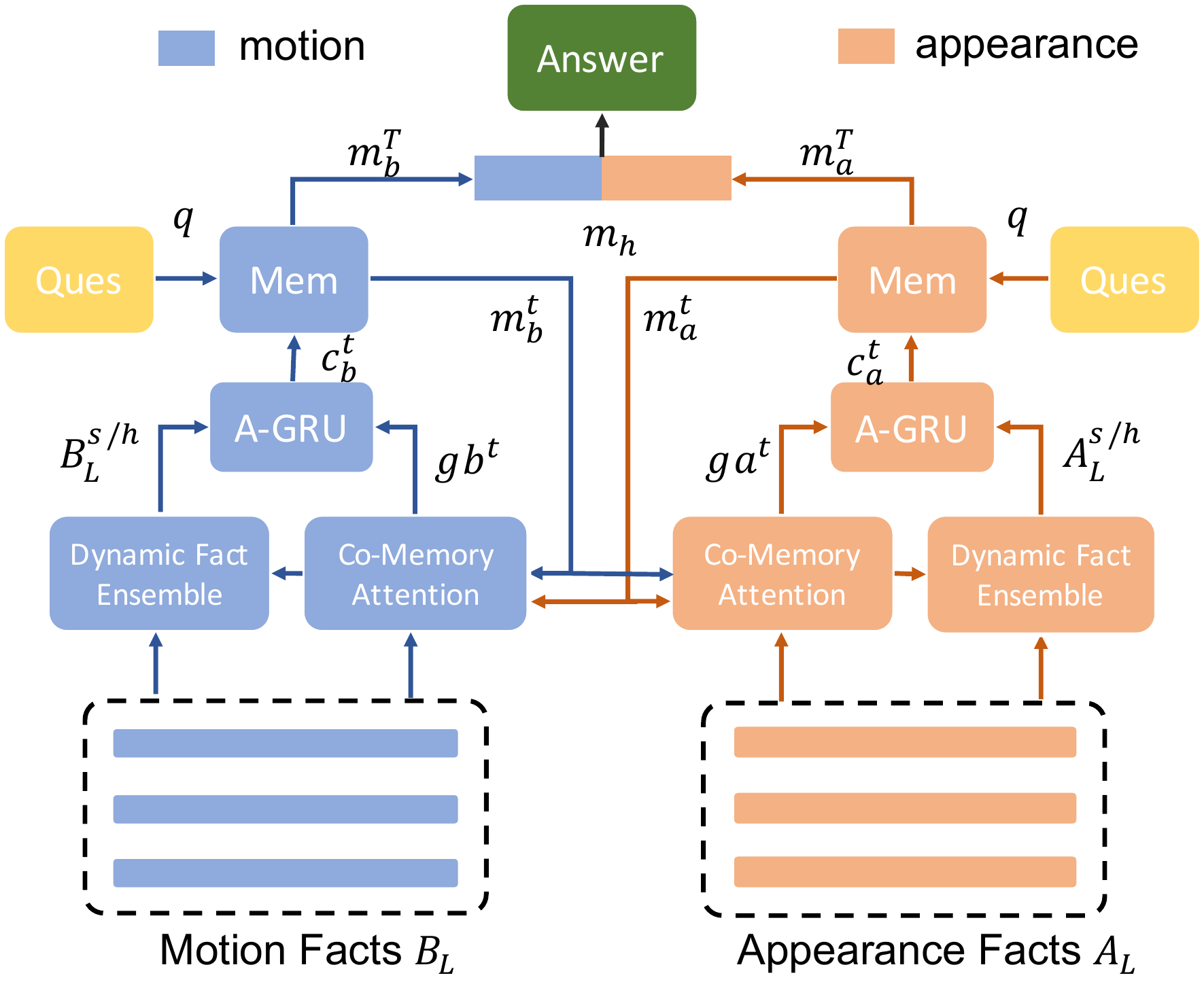}
    \caption{Co-memory attention module extracts useful cues from both appearance and motion memories to generate attention $ga^t$/$gb^t$ for motion and appearance separately. Dynamic fact ensemble takes the multi-layer contextual facts $\mathbf A_L$/$\mathbf B_L$ and the attention scores $ga^t$/$gb^t$ to construct proper facts $A_L^{s/h}$/$B_L^{s/h}$, which are encoded by an attention-based GRU. The final hidden state $c_b^t$/$c_a^t$ of the GRU is used to update the memory $m_b^t$/$m_a^t$. The final output memory $m_h$ is the concatenation of the motion and appearance memory, and it is used to generate answers. }
      \label{fig:comemory}
\end{figure}
\begin{align}
\begin{split}
    za_{i,j}^t&=\mbox{tanh}\left (\mathbf W_a^2 \left ( a^j_i+\mathbf W_a^1 [m_a^{t-1},q] \right ) \right ) \\
    ga_{i,j}^t&=\mathbf W_a^4 \left ( za_{i,j}^t+\mathbf W_a^3 [m_b^{t-1}, q] \right ) 
\end{split}
\end{align} 
\begin{align}
\begin{split}
    zb_{i,j}^t&=\mbox{tanh}\left (\mathbf W_b^2 \left ( b^j_i+\mathbf W_b^1 [m_b^{t-1},q] \right ) \right ) \\
    gb_{i,j}^t&=\mathbf W_b^4 \left ( zb_{i,j}^t+\mathbf W_b^3 [m_a^{t-1}, q] \right )
\end{split}
\end{align} 
where $\mathbf W_a^1$, $\mathbf W_a^2$, $\mathbf W_a^3$, $\mathbf W_a^4$, $\mathbf W_b^1$, $\mathbf W_b^2$, $\mathbf W_b^3$ and $\mathbf W_b^4$ are weight parameters. $ga_{i,j}^t$ and $gb_{i,j}^t$ are attentions used in dynamic fact ensemble and memory update.

\begin{figure}[]
  \centering
    \includegraphics[width=0.25\textwidth]{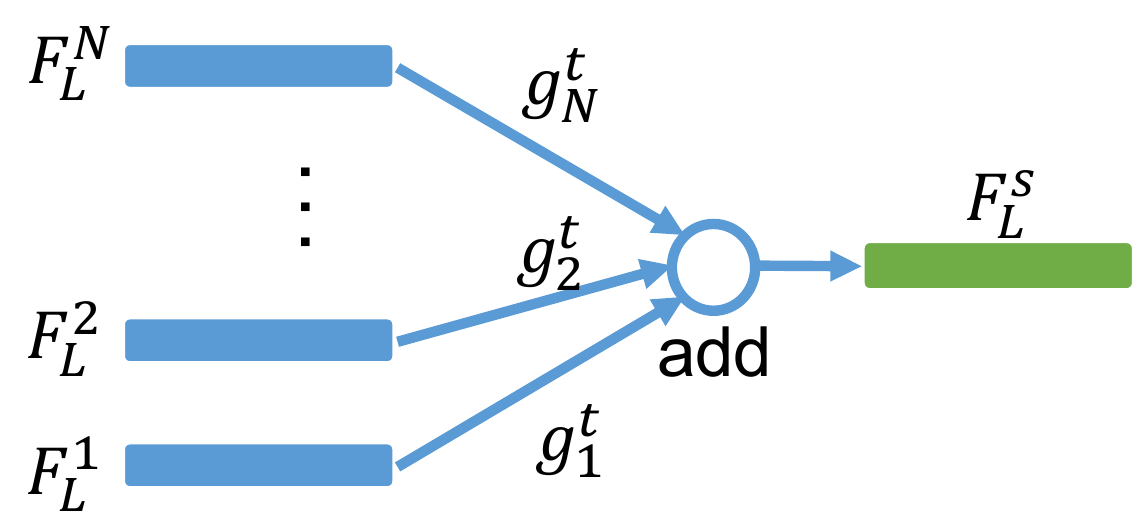}
    \caption{Multi-layer contextual facts are dynamically constructed via a soft attention fusion process, which computes a weighted average facts according to the attention. }
      \label{fig:ensemble}
\end{figure}

\textbf{Dynamic fact ensemble.}
As shown in Section \ref{sec:facts}, we build a multi-layer contextual facts set $\mathbf F_L=\{F^1_{L}, F^2_{L},..., F^N_{L}\}$ for motion and appearance separately, which have the same temporal resolution, but represent different contextual information. There are two reasons that the facts should be selected dynamically: (1) Different types of questions may require different level of representations, \eg the ``bulldog color'' and the ``cat lick'' questions given in Section \ref{sec:intro}; (2) During the multiple cycles of the fact reading, each cycle may focus on different level of information. We designed an attention-based fact ensemble methods shown in Figure \ref{fig:ensemble}. 
For simplicity, we use $g_{i,j}^t$ to represent the attention gate, which is actually $ga_{i,j}^t$ for appearance and $gb_{i,j}^t$ for motion. We calculate Softmax over $g_{i,j}^t$ along level axis (\ie $i$)  to get attention scores $s_{i,j}^t$. 


The ensemble facts can be represented as  
\begin{equation}
    F_{t}^s:\{f_{j}^t=\sum_{i=0}^N s_{i,j}^t f_{j}^i\}_{j=1}^L 
\end{equation} 
where $f_{i,j}$ is the fact vector of level $i$ and step $j$ in the contextual facts $\mathbf F_L$. The attention scores used in the later fact encoding process are given by
\begin{equation}
    s_{j}^t=\mbox{softmax}(\frac{1}{N}\sum_{i=0}^N g_{i,j}^t), j=1,2,...,L
\end{equation}
where the Softmax is computed along $j$ axis.
 
\textbf{Memory update.} The fact encoding processes are conducted separately for motion and appearance, which adopts an attention based GRU \cite{xiong2016dynamic} to generate contextual vectors $c_a^t$ and $c_b^t$ for appearance and motion in the $t$-th cycle. Motion memory $m_b^t$ and appearance memory $m_a^t$ are updated separately as follows.

\begin{equation}
    m_a^t=\mbox{FC}([m_a^{t-1},q,c_a^t])
\end{equation}

\begin{equation}
    m_b^t=\mbox{FC}([m_b^{t-1},q,c_b^t])
\end{equation}
where $\mbox{FC}$ means fully-connected layer, ReLU is used as the non-linear activation. The final output memory $m_h$ is the concatenation of $m_a^T$ and $m_b^T$, where $T$ is the number of cycles.

\subsection{Answer Module}
Following \cite{Jang_2017_CVPR}, we model the four tasks in TGIF-QA \cite{Jang_2017_CVPR} into three different types: multiple-choice, open-ended numbers and open-ended words. 

For multiple-choice, we use a linear regression function that takes the memory state $m_h$ and outputs a real-valued score for each answer candidate.
\begin{equation}
    s=\mathbf W_m^T m_h
\end{equation}
where $\mathbf W_m$ are weight parameters. The model is optimized by hinge loss between the scores for correct answers $s_p$ and the scores for incorrect answers $s_n$, $\max(0, 1 + s_{n} - s_{p})$. This decoder is used to solve repeating action and state transition tasks.

For open-ended numbers, we also use a linear regression function which takes the memory state $m_h$ and outputs an integer-valued answer.
\begin{equation}
    s=[\mathbf W_n^T m_h+b]
\end{equation}
where $[.]$ means rounding. We adopt $\ell_2$ loss between the groundtruth value and the predicted value to train the model, which is used to solve the repetition count task.

For open-ended words, we treat this as a classification problem. A linear function that takes the final memory state $m_h$ followed by a softmax layer is adopted to generate answers.
\begin{equation}
\label{eq:decoder_oe}
  \mathbf o = \mbox{softmax} \left ( \mathbf  W_w^{\top} \mathbf m_h + \mathbf b \right )
\end{equation}
where $\mathbf W_w$ are weight parameters and $\mathbf b$ is bias. Cross-entropy loss is used to train the model and this type of decoder is used in Frame QA task.

For each task, we train a separate model by the answer decoder and loss mentioned above. The model of each task is trained and evaluated individually.

\section{Evaluation}
\label{sec:eval}
In this section, we describe the dataset and evaluation settings, and discuss the experiment results.

\subsection{Dataset}
We evaluate the proposed model on TGIF-QA dataset~\cite{Jang_2017_CVPR}, which is a large-scale dataset introduced by Jang \etal for Video QA. The dataset consists of 165k QA pairs collocted from 71k animated Tumblr GIFs. There are four types of tasks: repetition count, repetition action, state transition and frame QA. First three tasks are unique to videos and require temporal reasoning to answer them.

\textbf{Tasks.} \textit{Repetition count} is an open-ended task to count the number of repetition of an action (\eg ``How many times does the cat
lick?"). There are 11 possible answers (\ie from 0 to 10+) in total. \textit{Repetition action} is a 5-option multiple choice task, which is asking about the name of the action that happened specific times (\eg ``what does the duck do 3 times?"). \textit{State transition} is also a 5-option multiple choice task which can be answered by understanding the transition of two states in a video (\eg ``What does the woman do after drink water?"). Besides, TGIF-QA also provides a traditional \textit{frame QA} task (\ie image QA). The image QA questions of previous datasets \cite{Antol_2015_ICCV, ren2015exploring, malinowski2014multi} can be answered by getting effective information from a single given image; but for frame QA in TGIF-QA dataset, the model needs to find the most relevant frame among all frames in the video to answer the question correctly. Frame QA is defined as an open-ended task. The number of QA pairs of TGIF-QA for the four tasks are shown in table \ref{tbl:dataset}.

\begin{table}[]
\centering
\caption{Number of samples of different tasks in TGIF-QA dataset.}
\vspace{5pt}
\label{tbl:dataset}
\begin{tabular}{l|cccc}
\hline
\# QA pairs     & Action    & Trans     & Count     & Frame     \\ \hline
Training        & 20,475    & 52,704    & 26,843    & 39,392    \\ \hline
Testing         & 2,274     & 6,232     & 3,554     & 13,691    \\ \hline
Total           & 22,749    & 58,936    & 30,397    & 53,083    \\ \hline
\end{tabular}
\end{table}

\textbf{Metric.} For the task of repetition count, the Mean Square Error (MSE) between the predicted count value and the groundtruth count value is used for evaluation. For repetition action, state transition and frame QA, classification accuracy (ACC) is used as the evaluation metric.

\subsection{Implementation Details}
\textbf{Appearance and motion features.} Since the frames per second (FPS) of the GIFs in TGIF-QA \cite{Jang_2017_CVPR} vary, we extract frames from all GIFs with the FPS that is specified by the corresponding GIF file. The long videos are cut into small units, each unit contains 6 frames.

To extract unit-level video features, we use ResNet-152 \cite{He_2016_CVPR_resnet}  to process the central frame of a unit, and the outputs of ``pool5'' layer ($\in{\mathbb {R}^{2,048}}$) of ResNet-152 is used as our appearance features. To utilize motion information, we extract optical flow inside a video unit, and use the flow CNN from two-stream model \cite{xiong2016cuhk} to get unit-level flow features. Specifically, the two-direction dense optical flows \cite{farneback2003two} which are calculated between two adjacent frames in a six-consecutive-frame unit are fed into the pre-trained flow CNN model, which is a BN-Inception network\cite{ioffe2015batch}. Then we take the feature map of the ``global\_pool'' layer ($\in{\mathbb {R}^{1,024}}$) as the raw optical flow features. Finally, we down-sample the feature dimension by average pooling and get a 2048-dimension vector as our two-direction optical flow feature. In this process, we pad the first or last frame if we didn't have enough frames centered at each step. We set the temporal resolution of video features to be 34, long feature sequences are cut and short one are padded.

\textbf{Contextual facts.} The output channel number of each layer in the conv-deconv networks is 1024, temporal conv filter size is 3 with stride 1, deconv layer with stride 2, max pool filter size is 2 with stride 2. We build $N=3$ layers of contextual facts.

\textbf{Co-memory module.} The size of memory state $m_a$ and $m_b$ is set to be 1024. The hidden state size of the GRU for fact encoding is 512. $za_{i,j}^t$ and $zb_{i,j}^t$ in equation (4) and (5) are 512-dimensional.

\textbf{Question and answer embedding.} For each word in the question, we use a pre-trained word embedding model \cite{pennington2014glove} to convert it to a 300-dimension vector. All words in the question are processed by a two-layer GRU, whose hidden state size is 512. The final hidden state is used as question embedding. For action transition and repeating action, the candidate answers are a sequence of words, thus we use the same method as the one for encoding questions to encode the answer.

\textbf{Training details.} We set the batch size to 64. Adam optimizer \cite{kingma2014adam} is used to optimize the model, the learning rate is set to 0.001. For each task, we train the model for 50 epochs.

\subsection{System Baselines}
Besides co-memory networks, there are two direct methods to make use of motion and appearance information: fact concatenation and memory concatenation, which are used as system baselines.

\textbf{Fact concatenation.} This baseline method simply concatenate the input motion facts and appearance facts, $\{b_{i}\}$ and $\{a_{i}\}$ along the feature dimension. The concatenated vector $\{h_{i}\}$ which is $d_b+d_a$ dimensional is used as input facts for multi-level contextual fact module. Only one memory module is used.

\textbf{Memory concatenation.} In this baseline method, instead of concatenating the input facts, we use two separate memory modules: one for appearance, the other for motion, and concatenate the final motion memory states $m_b^T$ and the final appearance memory states $m_a^T$ to $m_f^t$ together, which are used to decode answers. Co-memory attention mechanism is not used in this baseline.

\subsection{Experiments on TGIF-QA}
We first evaluate the co-memory attention module by comparing it with the two baseline method ``fact concatenation" and ``memory concatenation". Second, we evaluate the multi-level contextual facts and the dynamic fact ensemble.  Finally, we compare our method with the previous state-of-the-art methods.

\textbf{Co-memory attention.} In this experiment, we set the layer of contextual facts to be 1, and dynamic fact ensemble is not used. The number of memory updates  $T=2$. We compare co-memory attention mechanism with ``fact concatenation" (fact-concat) and ``memory concatenation" (mem-concat) to see the effectiveness of co-memory attention , the results are shown in Table \ref{tbl:comem}.
\begin{table}[h]
\centering
\caption{Evaluation of co-memory attention mechanism on TGIF-QA. ``Action" is repetition action (ACC \%), ``Trans" is state transition (ACC \%), ``Count" is repetition count (MSE) and ``Frame" is frame QA (ACC \%). }
\vspace{5pt}
\label{tbl:comem}
\begin{tabular}{l|cccc}
\hline
Method      & Action & Trans & Count & Frame \\ \hline
Fact-concat &    65.0    &  71.2     &    4.34   &     49.9  \\ \hline
Mem-concat  &   64.5     &  70.7    &   4.39    &    50.2   \\ \hline
Co-memory   &    \textbf{66.8} & \textbf{73.2}   &  \textbf{4.21}    &    \textbf{51.0}   \\ \hline
\end{tabular}
\end{table}
We can see that co-memory attention outperforms fact-concat and mem-concat in all four tasks, which shows the effectiveness of the co-memory attention mechanism. We believe the reason is that co-memory attention exploits the knowledge that motion and appearance provide useful cues to each other in attention generation.

\textbf{Contextual facts and dynamic fact ensemble.} Dynamic fact ensemble collaborates with multi-level contextual facts to construct proper temporal fact representation, so we test them together. We build 3 layers of contextual facts and do experiments to test dynamic fact ensemble module. We use ``fact concatenation" as the top memory network. The results are shown in Table \ref{tbl:facts}: ``w/o ensemble" means that we don't build the multi-level contextual facts, but just use a single temporal conv layer (filter size is 1) to convert appearance and motion features into 1024-dimension vectors, which are used as input facts. 

\begin{table}[h]
\centering
\caption{ Evaluation of dynamic fact ensemble on TGIF-QA. ``Action" is repetition action (ACC \%), ``Trans" is state transition (ACC \%), ``Count" is repetition count (MSE) and ``Frame" is frame QA (ACC \%). }
\vspace{5pt}
\label{tbl:facts}
\begin{tabular}{l|cccc}
\hline
Method      & Action & Trans & Count & Frame \\ \hline
w/o ensemble &   65.0    &  71.2     &    4.34   &     49.9       \\ \hline
w/ ensemble   &    66.3    &    72.5   &  4.30     &  50.4     \\ \hline
\end{tabular}
\end{table}

It can be seen that the ensemble provides better results. 
We believe the reason is that the attention-based fact fusion optimizes the ensemble process by using weighted average of the contextual facts, and avoids just using only one of them, which may make the facts sub-optimal.

\textbf{How many cycles of memory update are sufficient?} We test the co-memory attention model with different memory update times $T=1,2,3$ to see how many cycles of memory update are sufficient for video QA task. The dynamic fact ensemble is not used in this experiment. The results are shown in Table \ref{tbl:time}.

\begin{table}[h]
\centering
\caption{ Comparison on cycles of memory update on TGIF-QA. ``Action" is repetition action (ACC \%), ``Trans" is state transition (ACC \%), ``Count" is repetition count (MSE) and ``Frame" is frame QA (ACC \%).}
\vspace{5pt}
\label{tbl:time}
\begin{tabular}{l|cccc}
\hline
Method& Action & Trans & Count & Frame \\ \hline
$T=1$   &    65.1    &     69.9  &   4.35    &   50.5     \\ \hline
$T=2$   &   66.8     &    73.2   &    4.21   &     51.0  \\ \hline
$T=3$   &   66.5       &    73.1   &    4.24   &     51.1  \\ \hline
\end{tabular}
\end{table}
We can see that two cycles ($T=2$) of memory update gives the best performance on the task of ``Action", ``Trans" and ``Count". For ``Frame", $T=2$ and $T=3$ have similar results. Comparing the results of $T=2$ and $T=1$ in ``Trans", we can see that $T=2$ improves the performance by 3.3\%, we believe the reason is that multiple cycles of fact reading and memory update allow the model to focus on different parts of the video in each cycle. The performance begins to saturate at $T=3$. 

\textbf{Comparison with state-of-the-art method.} There are two version of TGIF-QA, we report the performance of the second version, which is released by the authors of \cite{Jang_2017_CVPR} on Arxiv. The first version is originally reported in the CVPR version of \cite{Jang_2017_CVPR}.
State-of-the-art method \cite{Jang_2017_CVPR} on TGIF-QA adopted a dual-LSTM based approach with both spatial and temporal attention. Originally, their model is trained on C3D \cite{Tran_2015_ICCV_c3d} temporal feature and ResNet-152 \cite{He_2016_CVPR_resnet} frame feature. However, our method adopts Flow CNN model (Inception) for motion and ResNet-152 for appearance. Thus, for fair comparison, we train their model (https://goo.gl/SVKTP9) with our features on all four tasks in TGIF-QA. The results are shown in Table \ref{tbl:sta}.   In Table \ref{tbl:sta}, ``SP" means spatial attention, ``TP" means temporal attention, ``(R+C)" means ResNet-152 features and C3D features, ``(R+F)" means ResNet-152 features and Flow CNN features (our feature). We also list methods ``VIS-LSTM" \cite{ren2015exploring} and ``VQA-MCB" \cite{fukui2016multimodal}, which are provided in \cite{Jang_2017_CVPR}.

\begin{table}[t]
\centering
\caption{Comparison with the state-of-the-art method on TGIF-QA dataset. ``Action" is repetition action (ACC \%), ``Trans" is state transition (ACC \%), ``Count" is repetition count (MSE) and ``Frame" is frame QA (ACC \%).}
\vspace{5pt}
\label{tbl:sta}
\setlength\tabcolsep{3.5pt} 
\begin{tabular}{l|cccc}
\hline
Model          & Action & Trans & Frame & Count \footnotemark \\ \hline
VIS+LSTM(aggr) \cite{ren2015exploring} & 46.8 & 56.9 & 34.6 & 5.09 \\ 
VIS+LSTM(avg) \cite{ren2015exploring} & 48.8 & 34.8 & 35.0 & 4.80 \\ \hline
VQA-MCB(aggr) \cite{fukui2016multimodal}  & 58.9 & 24.3 & 25.7 & 5.17 \\ 
VQA-MCB(avg) \cite{fukui2016multimodal}  & 29.1 &  33.0 &  15.5 & 5.54 \\ \hline
Yu \etal \cite{Yu_2017_CVPR}  &  56.1 & 64.0 & 39.6  & 5.13 \\ \hline
ST(R+C) \cite{Jang_2017_CVPR} & 60.1 &65.7  & 48.2 & 4.38 \\ 
ST-SP(R+C) \cite{Jang_2017_CVPR}& 57.3 & 63.7 & 45.5 & 4.28 \\ 
ST-SP-TP(R+C) \cite{Jang_2017_CVPR}& 57.0 & 59.6 & 47.8 & 4.56 \\ 
ST-TP(R+C) \cite{Jang_2017_CVPR}& 60.8 & 67.1 & 49.3 & 4.40 \\ 
ST-TP(R+F) & 62.9 & 69.4 & 49.5 &  4.32\\ \hline
Co-memory (w/o DFE) &  66.8 & 73.2 & 51.0 &  4.21\\
Co-memory (full)&  \textbf{68.2} & \textbf{74.3} & \textbf{51.5} & \textbf{4.10}\\ 

\hline
\end{tabular}
\end{table}
\footnotetext{We found an evaluation mistake in \cite{Jang_2017_CVPR} (https://goo.gl/SVKTP9) on count task. The new performances updated by the authors are listed here.}

\begin{figure*}[h]
  \centering
    \includegraphics[width=0.96\textwidth]{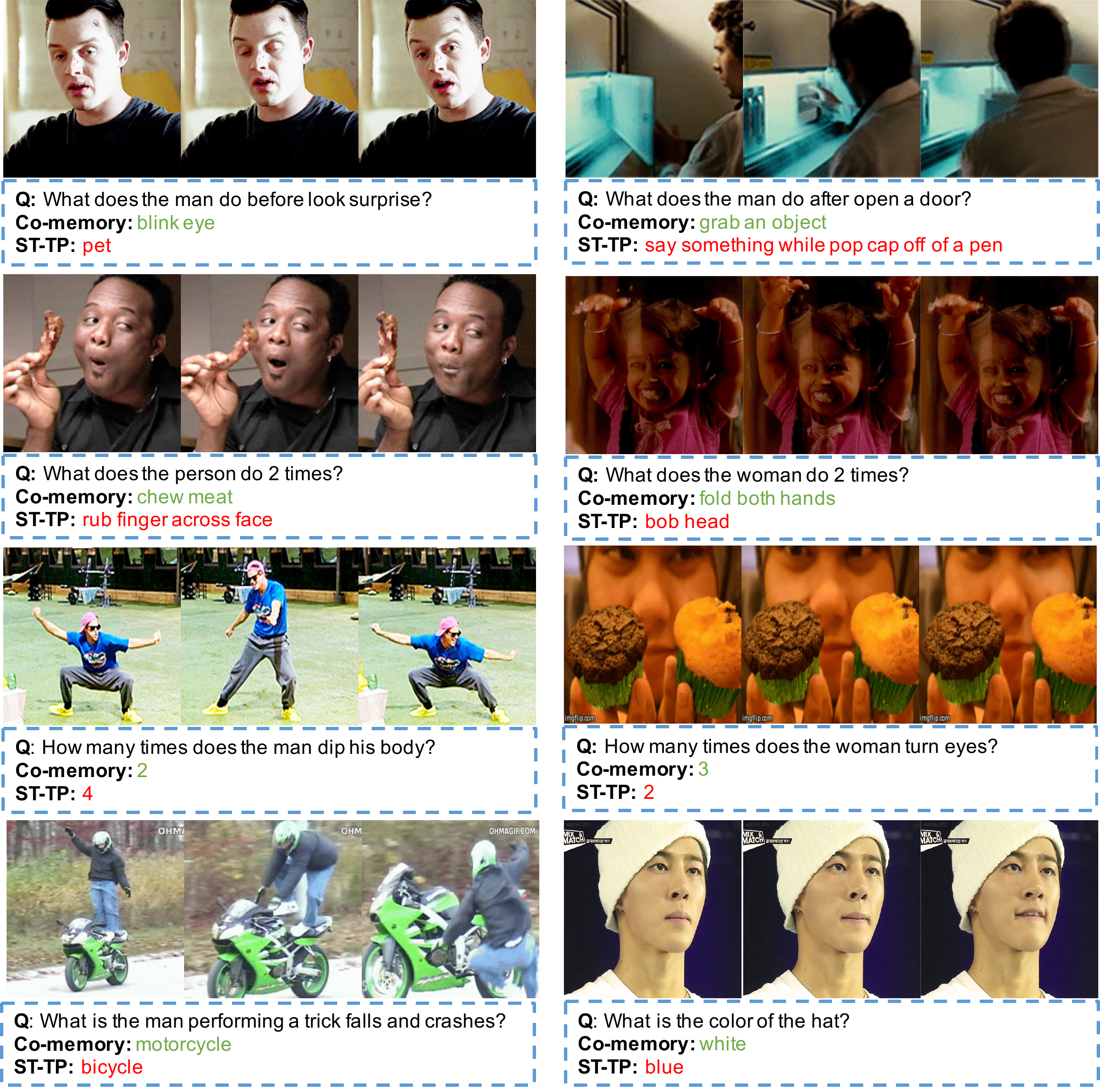}
    \caption{Examples on state transition, repetition action, repetition count and frame QA are shown in 1st, 2nd, 3rd and 4th row. ST-TP is the temporal attention model from \cite{Jang_2017_CVPR}. Green is for correct prediction and red is for wrong prediction.}
      \label{fig:example-final}
\end{figure*}

There are two co-memory variants shown in Table \ref{tbl:sta}: ``co-memory (w/o DFE)" uses co-memory attention with $T=2$ memory update, but not dynamic fact ensemble; ``co-memory (full)" uses co-memory attention with $T=2$ memory update and dynamic fact ensemble (soft fusion) on 3-layer contextual facts. We can see that our method outperforms the state-of-the-art method significantly on all four tasks. Some visualization examples are shown in Figure \ref{fig:example-final}.

\section{Conclusion}
Comparing with image QA, video QA deals with long sequences of images, which contains richer information in both quantity and variety. In addition, motion and appearance information are both important for video analysis, and usually correlated with each other and able to provide useful attention cues to the other. Motivated by these observations, we propose a motion-appearance co-memory network for video QA. Specifically, we design a co-memory attention mechanism that utilizes cues from both motion and appearance to generate attention,  a temporal conv-deconv network to generate multi-level contextual facts, and a dynamic fact ensemble method to construct temporal representation dynamically for different questions. We evaluate our method on TGIF-QA dataset, and outperforms state-of-the-art performance significantly.

\textbf{Acknowledgements.} This research was supported, in part, by the Office of Naval Research under grant N00014-18-1-2050.

{\small
\bibliographystyle{ieee}
\bibliography{egbib}
}

\end{document}